\documentclass[lettersize,journal]{IEEEtran}
\usepackage{amsmath,amsfonts}
\usepackage{algorithmic}
\usepackage{algorithm}
\usepackage{array}
\usepackage{textcomp}
\usepackage{stfloats}
\usepackage{url}
\usepackage{verbatim}
\usepackage{graphicx}
\usepackage{cite}

\usepackage{color}
\usepackage{booktabs}
\usepackage{subfigure}
\usepackage{ragged2e}
\usepackage{textcomp}
\usepackage{multirow}

\usepackage{amsmath,amssymb, bm}
\newtheorem{definition}{Definition}
\newtheorem{theorem}{Theorem}

\newtheorem{proof}{Proof}

\usepackage{listings}

\begin{document}

\title{UACER: An Uncertainty-Adaptive Critic Ensemble Framework for\\ Robust Adversarial Reinforcement Learning}

\author{
        Jiaxi Wu, Tiantian Zhang, Yuxing Wang, Yongzhe Chang, Xueqian Wang
% \thanks{
%         This work was supported by the China Postdoctoral Science Foundation under Grant No. 2025M781490, the Guangdong Provincial Natural Science Foundation under Grant No. 2024A1515010003, the Shenzhen Natural Science Foundation under Grant No. JCYJ20240813112007010 and Cross-disciplinary Fund for Research and Innovation of Tsinghua SIGS under Grant No. JC2024002. ({\em Corresponding authors: Tiantian Zhang; Xueqian Wang.})} 
\thanks{Jiaxi Wu and Tiantian Zhang contributed equally to this work and are with the Shenzhen International Graduate School, Tsinghua University, Shenzhen 518055, China (e-mail: jx-wu24@mails.tsinghua.edu.cn; zhang.tt@sz.tsinghua.edu.cn).}
\thanks{Yuxing Wang, Yongzhe Chang and Xueqian Wang are with the Shenzhen International Graduate School, Tsinghua University, Shenzhen 518055, China (e-mail: wyx20@mails.tsinghua.edu.cn; changyongzhe@sz.tsinghua.edu.cn; wang.xq@sz.tsinghua.edu.cn).}
}% <-this % stops an unwanted space


\IEEEpubid{0000--0000/00\$00.00~\copyright~2021 IEEE}

\maketitle

\begin{abstract}
Robust adversarial reinforcement learning has emerged as an effective paradigm for training agents to handle uncertain disturbance in real environments, with critical applications in sequential decision-making domains such as autonomous driving and robotic control.
Within this paradigm, agent training is typically formulated as a zero-sum Markov game between a protagonist and an adversary to enhance policy robustness. However, the trainable nature of the adversary inevitably induces non-stationarity in the learning dynamics, leading to exacerbated training instability and convergence difficulties, particularly in high-dimensional complex environments.
In this paper, we propose a novel approach, Uncertainty-Adaptive Critic Ensemble for robust adversarial Reinforcement learning (UACER), which consists of two components: 
1) Diversified critic ensemble: A diverse set of $K$ critic networks is employed in parallel to stabilize Q-value estimation in robust adversarial reinforcement learning, reducing variance and enhancing robustness compared to conventional single-critic designs. 
2) Time-varying Decay Uncertainty (TDU) mechanism: Moving beyond simple linear combinations, we propose a variance-derived Q-value aggregation strategy that explicitly incorporates epistemic uncertainty to adaptively regulate the exploration-exploitation trade-off while stabilizing the training process.
Comprehensive experiments across several challenging MuJoCo control problems validate the superior effectiveness of UACER, outperforming state-of-the-art methods in terms of overall performance, stability, and efficiency.
\end{abstract}


\section{Introduction}
\IEEEPARstart{D}{eep} reinforcement learning (DRL) has established itself as a transformative paradigm for sequential decision-making, demonstrating remarkable success beyond gaming domains. Its applications now span real-world scenarios including precision robotic control \cite{huang2022reward, cui2022reinforcement,  DBLP:conf/iros/ValenciaWXGLM24, tang2025deep}, adaptive healthcare systems \cite{yu2021reinforcement,shirley2024reinforcement,pmlr-v235-luo24f}, and autonomous vehicle navigation \cite{wu2021learn,feng2023dense,zhou2024indoor}. 
However, a critical gap persists: most DRL policies are trained in idealized simulated environments that fail to capture the complexity of physical world deployment. This issue becomes particularly pronounced in certain high-dimensional complex tasks such as robotic control. Even minor discrepancies in physical parameters, such as friction, mass distribution, or object appearance, can lead to significant performance deterioration when policies are deployed to real-world systems. More specifically, these policies frequently overfit to simulation-specific dynamics, resulting in catastrophic performance declines when confronted with inevitable real-world disturbances \cite{gleave2019adversarial} or distribution shifts \cite{chen2021context,zhang2024dynamics}.
Addressing this vulnerability, recent research \cite{pinto2017robust,NEURIPS2020_f0eb6568,DBLP:conf/iclr/ReddiT0CD24} has prioritized the development of RL frameworks capable of counteracting adversarial disturbances and environmental variations.

\IEEEpubidadjcol

Robust adversarial reinforcement learning (RARL) \cite{pinto2017robust,DBLP:conf/ijcnn/ShengZDKCZ22,DBLP:conf/iclr/ReddiT0CD24} is a prominent class of methods in this research domain, specifically designed to enhance the stability and generalization capabilities of RL agents operating in uncertain or adversarially perturbed environments. 
The framework reformulates conventional RL as a two-player zero-sum Markov game \cite{littman1994markov, perolat2015approximate}, comprising a protagonist agent that optimizes policy parameters to maximize cumulative returns and an adversary agent that systematically generates worst-case adversarial perturbations to minimize these returns. 
Whereas this minimax formulation theoretically enables agents to develop resilience against worst-case perturbations, its practical deployment poses significant methodological challenges \cite{NEURIPS2020_fb2e2032}:
\emph{(i) Non-Stationary Training Dynamics:} The adversary's evolving policy induces continuous perturbations, dynamically altering the protagonist's observed state distribution and reward signals. This interaction creates a highly non-stationary learning environment where the protagonist's optimization target shifts iteratively. As a result, policy updates exhibit high variance, frequently destabilizing the training process.
\emph{(ii) Convergence Impediments:} The cyclical adversarial training regime forces the protagonist to adapt to an ever-changing adversary policy before its own policy converges. Without a fixed optimization target, the protagonist remains trapped in a persistent reactive adaptation loop, leading to significantly prolonged convergence times due to continual policy adjustments, or even completed convergence failure in cases of extreme non-stationary conditions.

To address the aforementioned stability and convergence problems, we propose \textbf{U}ncertainty-\textbf{A}daptive \textbf{C}ritic \textbf{E}nsemble robust adversarial \textbf{R}einforcement learning (UACER), an effective general framework that advances conventional robust adversarial RL by systematically integrating critic ensemble methodology with uncertainty-adaptive policy optimization. This integration significantly improves both training stability and convergence speed in adversarial environments.
The technical foundation of UACER addresses a fundamental limitation of single-critic architectures, which are particularly vulnerable to adversarial perturbations and distributional outliers, leading to unstable Q-value estimation \cite{NEURIPS2021_3d3d286a,peer2021ensemble}. To overcome this limitation, UACER employs a parallel ensemble of critic networks, yielding three key benefits: 
Firstly, by aggregating Q-value estimates from multiple critics, the variance in Q-function estimation and policy gradient updates is reduced, thereby enhancing the stability of training;
Secondly, the diversified value function approximation estimators effectively suppress training oscillations and improve robustness against adversarial disturbances;
Thirdly, the resulting improvement in stability and robustness facilitates faster policy convergence during iterative adversarial interactions.
For the final Q-value determination session, we present a Time-varying Decay Uncertainty (TDU) Q-value aggregation function to adaptively combine outputs from multiple critic networks for policy optimization. The TDU mechanism is designed to: 
(i) encourage extensive exploration during the initial training phases by maintaining a higher tolerance for optimistic uncertainty, and
(ii) progressively transition toward more conservative value estimation in later stages to ensure stable convergence.
This approach allows for a dynamically adaptive balance between exploration and exploitation, while also preserving resilience against adversarial perturbations throughout the learning process.

The contributions of this article are summarized as follows.
\begin{itemize}
    \item[1)] We propose a novel critic ensemble method for robust adversarial reinforcement learning, called UACER, which achieves more stable and accurate Q-value estimation by employing multiple base critic predictors.  
    
    \item[2)] We introduce the TDU function, a principled decision strategy for adaptively combining Q-value predictions from multiple critics for policy optimization. This mechanism dynamically balances exploration and exploitation while ensuring stable policy updates throughout the training process. 
    
    \item[3)] Quantitative experimental results on several complex MuJoCo tasks demonstrate that UACER consistently outperforms state-of-the-art methods, including QARL and related baselines, in terms of overall performance, stability, and efficiency.
\end{itemize}

\section{Related Work}
\subsection{Robust Adversarial Reinforcement Learning}
The concept of adversarial training for robust RL was first established by RARL \cite{pinto2017robust}, which formulated the problem as a zero-sum Markov game between a protagonist agent and an adversarial opponent. This paradigm demonstrated that exposing agents to strategically generated perturbations during training could significantly enhance their robustness against real-world distributional shifts.
Building upon this framework, subsequent research has developed specialized improvements targeting several key limitations.
For instance, RARARL \cite{DBLP:conf/icra/PanSGC19} incorporated explicit risk modeling to reduce catastrophic failures, while NR-MDP \cite{pmlr-v97-tessler19a} expanded the adversary's action space to generate more diverse and challenging disturbances.
Further optimization advances introduced subsequently by MixedNE-LD \cite{DBLP:conf/nips/KamalarubanHHRS20} and Cen et al. \cite{DBLP:conf/nips/CenWC21} employed stochastic gradient Langevin dynamics and extragradient methods, respectively, to improve the robustness of adversarial training.
More recently, A2P \cite{liu2024robust} proposed dynamically adjusting the intensity of action-space perturbations based on the agent's current relative performance, enabling more adaptive adversarial training.
Although these approaches have progressively addressed specific robustness challenges, they remain limited in handling the fundamental stability and convergence inherent in adversarial training. Our work advances this line of research by developing a unified framework that simultaneously improves training stability and convergence properties, while preserving and further enhancing the robustness benefits of adversarial training paradigms.

\subsection{Training Stability and Convergence Advances in RARL}
While RARL has markedly advanced the robustness of RL policies, its adversarial formulation introduces fundamental challenges in training stability and convergence. These limitations are not confined to complex settings but are inherent to the framework itself, as rigorously demonstrated by \cite{NEURIPS2020_fb2e2032}, who showed their persistence even in simple linear quadratic regulation problems.
In response, recent research has primarily evolved along three directions: 
(i) Constrained optimization approaches, where DI-CARL \cite{DBLP:conf/aaai/ZhaiLDZWY22} established stability guarantees through dissipation inequation constraint and $L_2$ gain performance in adversary RL, and Yu et al. \cite{yu2021robust} proposed an accelerated convergence method for linear quadratic games by employing Bregman divergence to effectively capture and adapt to the global structure of the constraint set.
(ii) Curriculum learning methods enhance stability and robustness by progressively exposing the agent to escalating adversarial difficulty, notably through CAT \cite{DBLP:conf/ijcnn/ShengZDKCZ22}, which employed a structured curriculum to systematically increase task complexity, thereby ensuring a stable and efficient training process.
(iii) Hybrid frameworks such as QARL \cite{DBLP:conf/iclr/ReddiT0CD24} that combine entropy regularization with quantal response equilibrium and use a temperature parameter to systematically manage adversary rationality in a curriculum learning manner, thereby accelerating convergence while easing the complexity of the saddle point optimization problem.
In contrast, our work adopts an ensemble-based perspective that provides inherent stability to adversarial RL architecture through diversified Q-value estimation. This approach maintains compatibility with existing optimization strategies and further enhances the stability and convergence properties.

\subsection{Q-Function Ensemble in Reinforcement Learning}
The integration of multiple Q-value estimators, known as Q-function ensemble, has become a widely adopted approach to enhance the stability and robustness of value-based deep RL. 
Early ensemble methods in the DQN family primarily address overestimation bias through averaged or pessimistic selection strategies, such as the simple averaging in Averaged DQN \cite{anschel2017averaged} and the min-Q operator used in TD3 \cite{fujimoto2018addressing}, Soft Actor-Critic (SAC) \cite{haarnoja2018soft}, and Maxmin Q-learning \cite{DBLP:conf/iclr/LanPFW20}. This conservative estimation approach was further extended by methods like MSG \cite{ghasemipour2022so} and EDAC \cite{an2021uncertainty}, which employed lower confidence bounds to systematically reduce overestimation.
Beyond bias reduction, Q-function ensemble also facilitates exploration. Bootstrapped DQN \cite{osband2016deep} promoted behavioral diversity via bootstrap sampling and multi-headed Q-networks, while UCB Exploration \cite{chen2017ucb} aggregated mean and variance from $K$ critics to guide action selection with an upper confidence bound. Similarly, UA-DDPG \cite{kanazawa2022distributional} supported flexible ensemble sizes and encouraged exploration by prioritizing actions with high epistemic uncertainty. 
To further mitigate instability and balance exploration-exploitation, SUNRISE \cite{lee2021sunrise} was proposed as a unified ensemble RL framework, which re-weighted target Q-values based on ensemble uncertainty and selected actions via UCB.
Despite empirical success in standard RL settings, these methods are primarily designed for stationary, non-adversarial environments, leaving their application in robust adversarial RL largely unexplored. Under adversarial disturbances, Q-value estimates exhibit significant fluctuation, and simplistic pessimistic value estimation can hinder policy learning. Unlike prior work, we introduce a diversified critic ensemble to stabilize Q-learning against strategic perturbations and propose a time-varying decay uncertainty mechanism for Q-value aggregation, tailored to promote rapid exploration and policy learning in non-stationary adversarial RL training.

\section{Preliminaries and Notations} 
This section introduces the core terminology used in our paper, covering standard RL concepts, the basic RL framework SAC, and the RARL paradigm formulated as a two-player zero-sum game. The notations employed in this article are summarized in Table \ref{notations}.

\subsection{Standard RL and Soft Actor-Critic}

\paragraph{Standard RL}
RL addresses the problem of sequential decision-making and is typically formulated as a Markov Decision Process (MDP), in which an agent interacts with an environment over discrete time steps.
At each time step $t$, the agent observes the current state $s_t$ and selects an action $a_t$ according to its policy $\pi$.
In response, the environment provides a reward $r_t$ and transitions the agent to the subsequent state $s_{t+1}$.
The objective of RL is to learn an optimal policy $\pi^{\ast}$ through repeated trial-and-error interactions and limited feedback, allowing the agent to effectively engage with the environment and maximize long-term cumulative rewards.

\paragraph{Soft Actor-Critic}
Similar to the family of existing robust adversarial RL methods \cite{pinto2017robust, DBLP:conf/nips/KamalarubanHHRS20, DBLP:conf/iclr/ReddiT0CD24}, we use SAC \cite{haarnoja2018soft} as the underlying RL algorithm in this paper. 
SAC is an off-policy actor-critic method that extends the standard RL objective through entropy regularization, thereby enhancing exploration and policy robustness. The algorithm optimizes a stochastic policy that maximizes the expected discounted return together with an entropy-based regularization term. 
Formally, let $\pi_{\phi}(s_t,a_t)$ represent the policy network parameterized by $\phi$, and $Q_{\theta}(s_t,a_t)$ denote the Q-function network with parameters $\theta$. The Q-function is updated by minimizing the soft Bellman residual:
\begin{equation}
\mathcal{L}_Q(\theta) = \mathbb{E}_{\tau_t \sim \mathcal{D}} \left[\big(Q_\theta(s_t, a_t)-R_t-\gamma V(s_{t+1})\big)^2\right],
\end{equation}
where $V(s_t)=\mathbb{E}_{a_t \sim \pi_{\phi}}\big[Q_{\theta}(s_t, a_t)-\alpha \log \pi_{\phi}(a_t|s_t)\big]$ is the soft state value function, $\tau_t$ is a transition, $\mathcal{D}$ is a replay buffer, and $\alpha$ is a temperature parameter. 
The policy is updated by minimizing the following objective:
\begin{equation}
\mathcal{L}_{\pi}(\phi)=\mathbb{E}_{s_t\sim\mathcal{D}}\Big[\mathbb{E}_{a_t\sim\pi_{\phi}}\big[\alpha\log\pi(a_t|s_t)-Q_{\theta}(s_t, a_t)\big]\Big].
\end{equation}
Collectively, these update rules establish a stable and sample-efficient learning foundation. In this work, we extend this framework to environments with uncertainty or adversarial disturbances through uncertainty-adaptive ensemble mechanisms.

\begin{table}[!t]
\caption{Notations and their descriptions.}
\label{notations}
\setlength{\tabcolsep}{0.5mm}
\renewcommand\arraystretch{1.2}
\begin{tabular}{ll}
  \toprule
  Notation                      & Description \\
  \midrule
  $\gamma$                         & discount factor \\
  $t$                              & time step \\
  $a_p,a_a$                        & actions taken by the protagonist and adversary \\
%   $\pi_p^\ast, \pi_a^\ast$         & approximate optimal policies of the protagonist and adversary \\
  $\theta_k^p, \theta_k^a$         & parameters of the $k^\text{th}$ critic networks for protagonist and adversary \\
  $\phi_p, \phi_a$                 & parameters of the actor networks for protagonist and adversary \\
  $\mathcal{D}_p, \mathcal{D}_a$   & replay buffer for protagonist and adversary \\
  $\pi_p, \pi_a$                   & policies of the protagonist and adversary \\
  $K$                              & number of ensemble critic networks \\
  $\alpha$                         & temperature parameter in SAC \\
  $\beta_0$                        & initial value of the time-decaying coefficient in TDU \\
  $\beta_{\text{min}}$             & a small constant for time-decaying coefficient in TDU \\
  $\lambda$                        & decaying speed in TDU \\
  $N$                              & total number of iterations \\
  \bottomrule
\end{tabular}
\end{table}

\subsection{RARL with Two-Player Zero-Sum Markov Games}
We consider an adversarial training framework for RL in adversarially perturbed environments, modeled as a two-player $\gamma$ discounted zero-sum Markov game \cite{littman1994markov, perolat2015approximate}: $\mathcal{M}=\langle \mathcal{S}, \mathcal{A}_p, \mathcal{A}_a, \mathcal{P}, \mathcal{R}, \gamma \rangle$, where $\mathcal{S}$ denotes the state space, $\mathcal{A}_p$ and $\mathcal{A}_a$ are the action spaces of the protagonist and adversary, respectively; $\mathcal{P}:\mathcal{S}\times\mathcal{A}_p\times\mathcal{A}_a\times\mathcal{S}\rightarrow\mathbb{R}$ defines the transition probability density, and $\mathcal{R}:\mathcal{S}\times\mathcal{A}_p\times\mathcal{A}_a\times\mathcal{S}\rightarrow\mathbb{R}$ is the reward of both players.
Our goal is to learn the policy of the protagonist (denoted by $\pi_p$) such that it can obtain maximum $\gamma$ discounted return in adversarial environments. Supposing the policy of the adversary is denoted by $\pi_a$, the optimization objective of the protagonist is to seek to maximize the following expected return:
\begin{equation}
\label{TZMG-goal}
\resizebox{.91\hsize}{!}{
$J_{\pi_p,\pi_a}(s)=\mathbb{E}_{a_p\sim\pi_p(\cdot|s), a_a\sim\pi_a(\cdot|s)}\!\left[\sum\limits_{t=0}^{\infty} \gamma^{t} r(s_t, a_p^t, a_a^t, s_{t+1}) \right],$}
\end{equation}
while the adversary simultaneously attempts to minimize it through its policy $\pi_a$.
The optimal solution corresponds to a minimax or Nash equilibrium satisfying:
\begin{equation}
\label{TZMG-optimal}
J_{\pi_p^\ast,\pi_a^\ast}(s)=\min_{\pi_a}\max_{\pi_p}J_{\pi_p,\pi_a}=\max_{\pi_p}\min_{\pi_a}J_{\pi_p,\pi_a}.
\end{equation}
In the practical implementation, the adversarial training follows an alternating procedure: at each iteration, we first hold the protagonist's policy fixed and update the adversary's policy to minimize the expected return $J$, then conversely freeze the adversary's policy and learn the protagonist's policy to maximize the return.
This procedure introduces training instability and convergence difficulties due to concurrent policy updates and adversarial perturbations. Our work specifically addresses these challenges through ensemble-based value estimation and uncertainty-adaptive policy optimization.

\section{Methodology}

\begin{figure*}[!t]
\centering
\includegraphics[width=.98\linewidth]{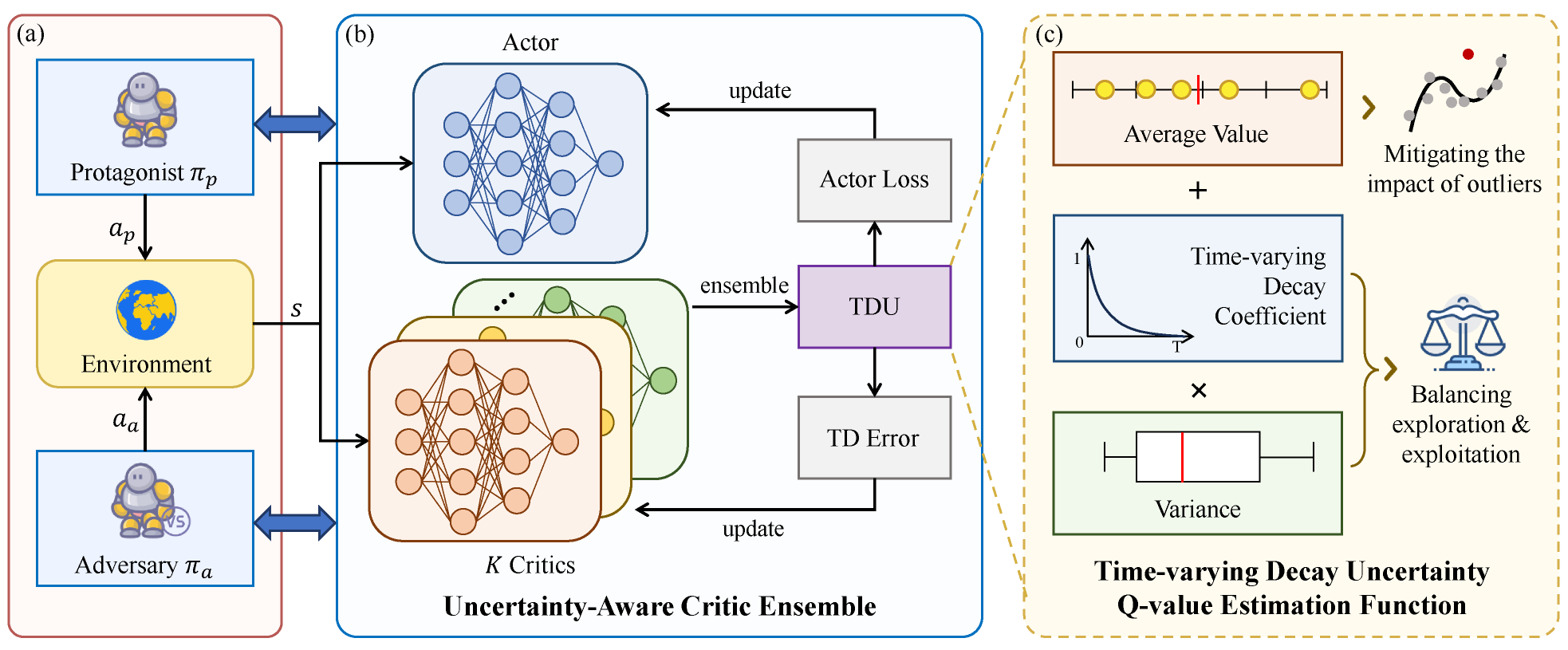} 
\caption{Schematic of the proposed UACER method, comprising: (a) robust adversarial RL components for protagonist-adversary interaction; (b) uncertainty-aware critic ensemble training architecture with $K$ parallel critics and TDU Q-value estimation; (c) detailed decomposition of TDU function.}
\label{UACER_framework}
\end{figure*}

In this section, we give a detailed description of UACER, whose overview is illustrated in Fig. \ref{UACER_framework}.
We first introduce the diversified critic ensemble framework for robust adversarial RL and then propose the variance-based TDU Q-value aggregation function, followed by theoretical guarantees for TDU's convergence properties and aggregation effectiveness.
In principle, our method can be used in conjunction with most modern robust adversarial RL algorithms, including recent advances like QARL \cite{DBLP:conf/iclr/ReddiT0CD24} as well as classical approaches such as RARL \cite{pinto2017robust}. 
For the sake of concreteness, we here present the QARL-based implementation as our primary instantiation.

\subsection{Diversified Critic Ensemble}

\begin{figure*}[!t]
    \centering
    \subfigure[\label{actor_ensemble}]
        {\includegraphics[width=0.28\linewidth]{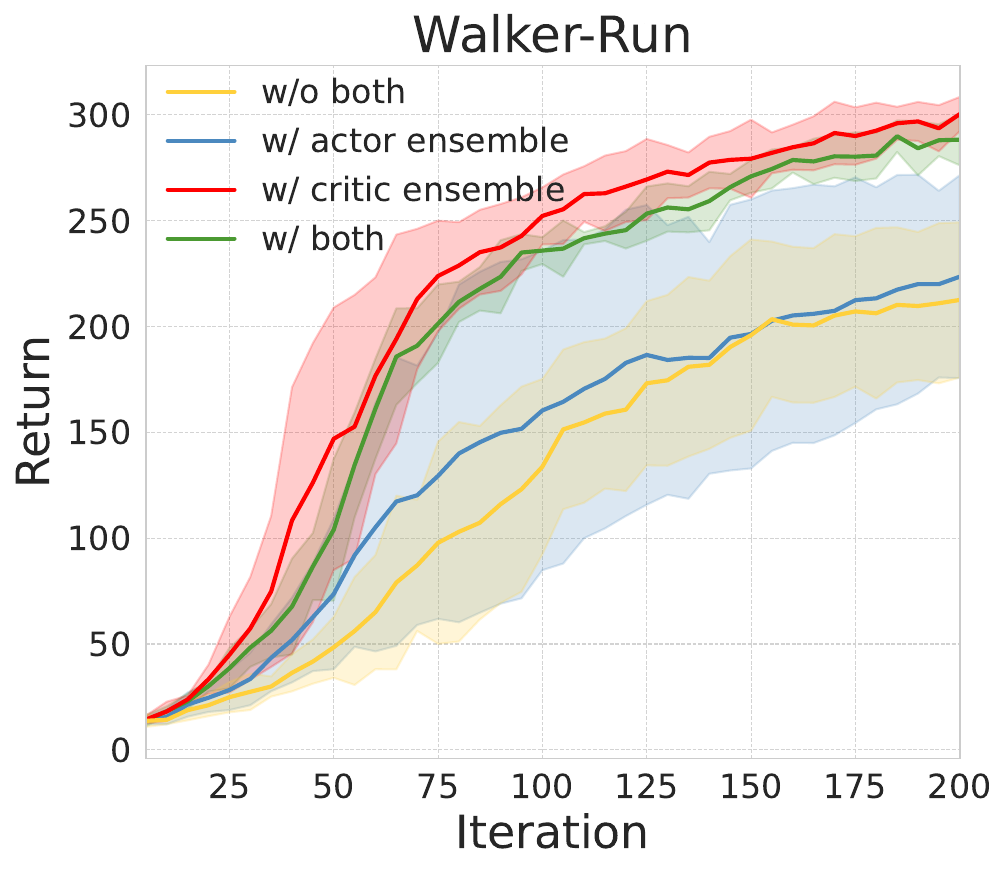}}
    \hspace{0.5cm}
    \subfigure[\label{diversity}]
        {\includegraphics[width=0.28\linewidth]{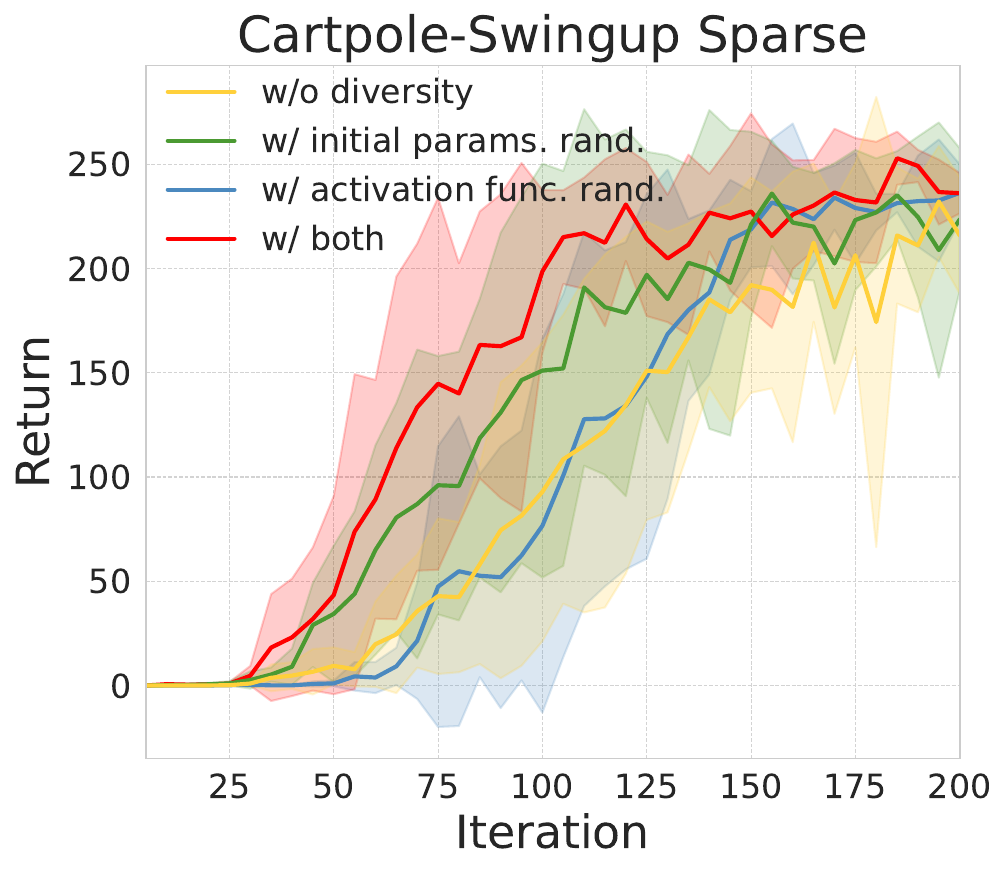}}
    \hspace{0.5cm}
    \subfigure[\label{coefficient}]
        {\includegraphics[width=0.28\linewidth]{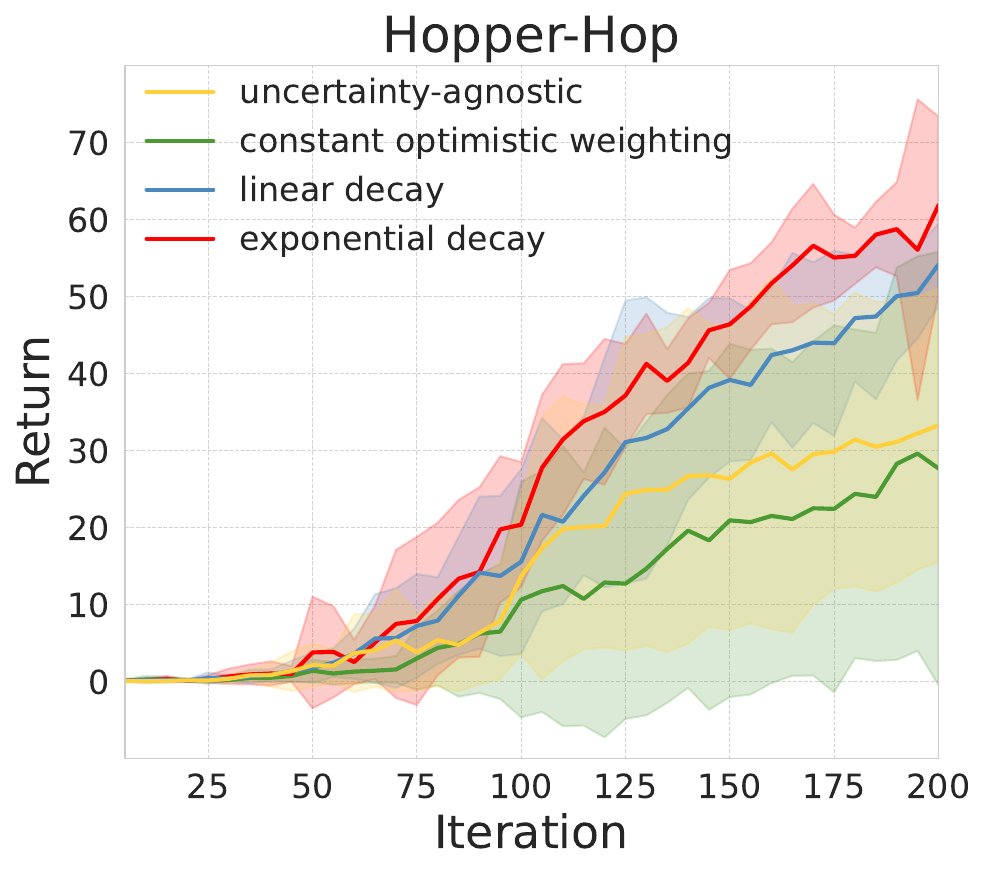}}
    \caption{Robustness evaluation curves of (a) ensemble component ablation, (b) diversity ablation, and (c) different uncertainty coefficients on MuJoCo swingup and locomotion tasks. Here and in related figures in the following, each curve shows the average return obtained for varying properties of the environment after the corresponding training iteration. The solid lines and shaded regions denote the average performance and the $95\%$ confidence intervals, respectively, across five seeds.}
\end{figure*}

\paragraph{The Unified Framework}
The UACER framework implements both protagonist and adversary as actor-critic agents based on the SAC algorithm. Each agent consists of a single policy network, denoted as $\pi_{\phi}$, and an ensemble of $K$ soft Q-function networks $\{Q_{\theta_k}\}_{k=1}^K$.
This design is motivated by the empirical insight, as shown in Fig. \ref{actor_ensemble}, that the critic ensemble significantly enhances the robustness of Q-value estimation, especially under adversarial perturbations. By evaluating Q-values through multiple independent critics, the framework mitigates the influence of localized noise or perturbations on any single estimator, thereby yielding more stable and reliable value estimates.
In contrast, although actor ensembles can improve sample efficiency and generalization \cite{lee2021sunrise, yang2022towards}, they often introduce additional training instability due to policy divergence among actors, which may produce conflicting behaviors that undermine the consistency and convergence of the learned policy.

\begin{algorithm}[!t]
    \caption{UACER}
    \label{alg:UACER}
    \renewcommand{\algorithmicrequire}{\textbf{Input:}}
    \renewcommand{\algorithmicensure}{\textbf{Output:}}
    
    \begin{algorithmic}[1]
        \REQUIRE Initial $K$ critics for protagonist: $\{Q_{\theta_k^p}\}_{k=1}^K$, $\{Q_{\bar{\theta}_k^p}\}_{k=1}^K$;\\
        \hspace{0.42cm} Initial $K$ critics for adversary: $\{Q_{\theta_k^a}\}_{k=1}^K$, $\{Q_{\bar{\theta}_k^a}\}_{k=1}^K$;\\
        \hspace{0.42cm} Initial actor $\pi_{\phi_p}$, $\pi_{\phi_a}$ for protagonist and adversary;\\
        \hspace{0.42cm} Replay buffer $\mathcal{D}_p=\emptyset$, $\mathcal{D}_a=\emptyset$;\\
        \hspace{0.42cm} Total number of iterations $N$;\\
        \hspace{0.42cm} TDU related parameters $\beta_0, \beta_{\min}, \lambda$.
        \ENSURE Protagonist's policy $\pi_{\phi_p}$.
        
        \FOR{each iteration $n\in\{1,\dots,N\}$}
            \STATE Update target critics for both agents:\\
            \hspace{0.4cm}
            $\{\bar{\theta}_k^p\}_{k=1}^K \leftarrow \{\theta_k^p\}_{k=1}^K$, $\{\bar{\theta}_k^a\}_{k=1}^K \leftarrow \{\theta_k^a\}_{k=1}^K$.
            \WHILE{performing the adversary update}
                \STATE Both agents interact with the environment and store the collected transitions in $\mathcal{D}_a$.
                \STATE Update $\{Q_{\theta_k^a}\}_{k=1}^K$ using Eqs.\eqref{critic_loss}, \eqref{soft_value} and \eqref{TDU_func}.
                \STATE Update $\pi_{\phi_a}$ using Eqs.\eqref{actor_loss} and \eqref{TDU_func}.
            \ENDWHILE
            \WHILE{performing the protagonist update}
                \STATE Both agents interact with the environment and store the collected transitions in $\mathcal{D}_p$.
                \STATE Update $\{Q_{\theta_k^p}\}_{k=1}^K$ using Eqs.\eqref{critic_loss}, \eqref{soft_value} and \eqref{TDU_func}.
                \STATE Update $\pi_{\phi_p}$ using Eqs.\eqref{actor_loss} and \eqref{TDU_func}.
            \ENDWHILE
        \ENDFOR
        \RETURN $\pi_{\phi_p}$.
    \end{algorithmic}
\end{algorithm}

During the adversarial training phase, both agents interact synchronously with the environment through their current policies, generating transition tuples $\tau_t=(s_t,a_{p_t},a_{a_t},r_t,s_{t+1})$ that are stored in the experience replay buffer $\mathcal{D}$. 
Given the symmetric nature of the protagonist-adversary learning paradigm, their parameter update principle can be uniformly formalized as follows:
At timestep $t$, the protagonist executes action $a_t=a_{p_t}$ and receives reward $R_t=r_t$, while the adversary selects action $a_t=a_{a_t}$ and obtains reward $R_t=-r_t$.
To reduce the variance in the Q-function estimate, we update parameter $\theta_k$ for each critic $Q_{\theta_k}(k\in\{1,2,\dots,K\})$ with the same target to minimize the soft Bellman residual:
\begin{equation}
\label{critic_loss}
    \mathcal{L}_{Q}(\theta_k)=\mathbb{E}_{\tau_t\sim\mathcal{D}}\Big[\big(Q_{\theta_k}(s_t,a_t) - R_t - \gamma V(s_{t+1})\big)^2\Big],
\end{equation}
with
\begin{equation}
   \label{soft_value}
    V(s_t)=\mathbb{E}_{a_t \sim \pi_{\phi}}\big[Q_{\bar{E}}(s_t, a_t)-\alpha \log \pi_{\phi}(a_t|s_t)\big],
\end{equation}
where $Q_{\bar{E}}(s_t, a_t)$ represents the ensemble Q-value estimate aggregated from $K$ target critics through our proposed TDU function, whose formal definition will be detailed in the next section.
Then, the policy $\pi_{\phi}$ can be updated by the ensemble Q-value estimate and minimizing: 
\begin{equation}
\label{actor_loss}
\mathcal{L}_{\pi}(\phi)=\mathbb{E}_{s_t\sim\mathcal{D}}\Big[\mathbb{E}_{a_t\sim\pi_{\phi}}\big[\alpha\log\pi(a_t|s_t)-Q_E(s_t, a_t)\big]\Big],
\end{equation}
where $Q_E(s_t,a_t)$ is the TDU output aggregated from $K$ critics.
The complete iterative optimization procedure of our unified framework is summarized in Algorithm \ref{alg:UACER}.

\paragraph{Diversity Enhancement} 
To encourage sufficient diversity among critics, we employ a dual randomization strategy combining parameter initialization and architectural design.
Each critic network undergoes independent Gaussian weight initialization with parameter noise injection, establishing distinct initial optimization paths \cite{plappert2018parameter}.
Additionally, we randomly assign activation functions (ReLU, LeakyReLU, ELU) \cite{dubey2022activation} to each layer during forward propagation.
This combined parametric and structural randomization systematically promotes representational diversity across critics while maintaining functional compatibility, with experimental results in Fig. \ref{diversity} empirically validating its effectiveness.

\subsection{Time-varying Decay Uncertainty Mechanism}

Conventional ensemble methods in RL typically stabilize training by averaging critic outputs, which has been proven effective in reducing variance of target approximation errors \cite{anschel2017averaged} and refining policy optimization directions \cite{chen2021randomized}.
Building on this foundation, we propose the TDU function, formally defined in Definition \ref{tdu_function}, as an optimistic aggregation mechanism tailored for $K$-critic ensembles in non-stationary environments.
The TDU mechanism explicitly considers epistemic uncertainty among critics through variance quantification and adaptively modulates its impact on Q-value estimation via time-decaying coefficients.
This dual mechanism design enables TDU to simultaneously enhance Q-value estimation stability and maintain an adaptive exploration-exploitation balance throughout the training process. 

\begin{definition}[TDU Function]
\label{tdu_function}
Consider a set of $K$ independent and identically distributed (i.i.d.) unbiased estimators $\{Q_{\theta_k}(s_t,a_t)\}_{k=1}^K$ for the true state-action value function $Q^{\ast}(s_t,a_t)$, where each estimator has finite variance. 
For the current adversarial training alternating iteration index $n\in\{1,2,...,N\}$ (where $N$ is the total number of iterations), the TDU function is defined as:
\begin{equation}
Q_E(s_t, a_t)=\mu_{Q}(s_t,a_t)+\beta(n) \cdot \sigma_{Q}(s_t,a_t),
\label{TDU_func}
\end{equation}
with
\begin{equation}
\mu_{Q}(s_t, a_t)=\frac{1}{K} \sum_{k=1}^{K} Q_{\theta_k}(s_t, a_t),
\end{equation}
\begin{equation}
\sigma_{Q}^{2}(s_t, a_t)=\frac{1}{K-1} \sum\limits_{k=1}^{K}\big[Q_{\theta_k}(s_{t}, a_t)-\mu_{Q}(s_t, a_t)\big]^{2},
\end{equation}
and
\begin{equation}
\beta(n) = \beta_0 \cdot e^{-\lambda n / N} + \beta_{\min}, \quad \text{s.t.} \quad \beta_0 + \beta_{\min} = 1
\label{time-varying_decay}
\end{equation}
is the coefficient to control the degree of considered uncertainty with time-varying decay, 
where $\lambda$ is a large positive constant, $\beta_0$ is the initial coefficient value, and $\beta_{\min}$ is a small positive value approaching zero.
\end{definition}

As evident from the above definition, the efficacy of the TDU function is principally governed by its time-varying decay coefficient $\beta(n)$, which adaptively modulates how epistemic uncertainty contributes to Q-value aggregation.
Fig. \ref{coefficient} compares the robustness performance of our proposed natural exponential decay formulation for $\beta(n)$ (given in Eq. \eqref{time-varying_decay}) with three baseline uncertainty-weighting strategies on the MuJoCo {\tt Hopper-Hop} task: 
(i) {\em uncertainty-agnostic}: $\beta(n)\equiv0$, equivalent to standard averaging;
(ii) {\em constant optimistic weighting}: $\beta(n)\equiv1$, fixed maximum uncertainty weighting;
(iii) {\em linear decay}: $\beta(n)$ decreasing linearly from $1$ to $0$.
The results demonstrate that time-decaying weighting strategies enable agents to learn significantly more robust policies compared with both uncertainty-agnostic and constant optimistic weighting approaches.
This advantage stems from their ability to adaptively transition from exploration-emphasizing behavior in early training stages to confidence-prioritizing predictions in later phases, thereby improving both estimation accuracy and learning efficiency.
Furthermore, our proposed natural exponential decay strategy achieves superior robustness performance and faster convergence than linear decay through its monotonically decreasing derivative, allowing faster reduction of the agent's uncertainty preference during early training while preserving essential exploration capacity for effective learning progression.

\subsection{Theoretical Guarantees for TDU}

In this section, we theoretically analyze the TDU mechanism through Theorem \ref{consisitency}, establishing its convergence guarantees as a Q-value estimator in adversarial RL settings.

\begin{theorem}[Unbiasedness and Consistency]
\label{consisitency}
Suppose there exist $K$ critics $\{Q_{\theta_k}(s_t, a_t)\}_{k=1}^K$ designed to estimate the true state-action value function $Q^{\ast}(s_t, a_t)$, each satisfying:
\begin{itemize}
    \item unbiased estimation:
        \begin{equation}
            E[Q_{\theta_k}(s_t, a_t)]=Q^{\ast}(s_t, a_t), \forall k\in\{1,\dots,K\};
        \end{equation}
    \item bounded variance:
        \begin{equation}
            \operatorname{Var}[Q_{\theta_k}(s_t, a_t)] \leq \sigma^{2}, \forall k\in\{1,\dots,K\}.
        \end{equation}
\end{itemize}
Then,
\begin{equation}
    \lim_{N \rightarrow \infty} \mathbb{E}\big[|Q_E(s_t, a_t)-Q^{\ast}(s_t, a_t)|\big] \leq \beta_{\min} \cdot \sigma.
\end{equation}
As $\beta_{\min} \to 0$, the ensemble estimator $Q_E(s_t, a_t)$ converges in probability to $Q^\ast(s_t, a_t)$:
\begin{equation}
Q_E(s_t, a_t) \overset{p}{\to} Q^{\ast}(s_t, a_t).
\end{equation}
\end{theorem}

UACER is built upon established robust adversarial RL frameworks, including RARL \cite{pinto2017robust}, MixedNE-LD \cite{DBLP:conf/nips/KamalarubanHHRS20}, and QARL \cite{DBLP:conf/iclr/ReddiT0CD24}, and thus inherits their theoretical guarantee of convergence to a Nash equilibrium under adversarial interactions. This foundation ensures stable and well-founded learning dynamics even in the presence of perturbations and strategic opponents.
Moreover, Theorem \ref{consisitency} formally establishes the asymptotic reliability of TDU’s Q-value estimation. It shows that as the number of training iterations grows sufficiently large and $\beta_{\min}$ approaches zero, the expected bias between TDU's Q-value estimates and the true Q-values diminishes asymptotically. 
This result confirms TDU's consistency as an aggregation method and demonstrates its theoretical capability to converge to the optimal value function given sufficient training.

The proof of Theorem \ref{consisitency} is provided as follows.

\begin{proof}
\label{proof_of_theorem1}
Given the unbiasedness condition $E[Q_{\theta_k}(s_t, a_t)]=Q^{\ast}(s_t, a_t)$ for all $k\in\{1,\dots,K\}$, it follows that the ensemble mean satisfies
\begin{equation}
\mathbb{E}\big[Q_{\theta_k}(s_t, a_t)\big] = \mu_Q(s_t, a_t) = Q^\ast(s_t, a_t).
\end{equation}
Consider the deviation of the ensemble estimate $Q_E(s_t, a_t)$ from the true value $Q^{\ast}(s_t, a_t)$:
\begin{equation}
\begin{aligned}
\big|Q_E(&s_t, a_t) - Q^\ast(s_t, a_t)\big| \\ 
& = \big|\mu_Q(s_t, a_t) - Q^\ast(s_t, a_t) + \beta(n) \cdot \sigma_Q(s_t, a_t)\big|.
\end{aligned}
\end{equation}
Then, applying the triangle inequality yields
\begin{equation}
\begin{aligned}
\big|Q_E(&s_t, a_t) - Q^\ast(s_t, a_t)\big| \\
& \leq \big|\mu_Q(s_t, a_t) - Q^\ast(s_t, a_t)\big| + \beta(n) \cdot \sigma_Q(s_t, a_t).
\end{aligned}
\end{equation}
Taking expectations on both sides and let the number of training iterations $N\to\infty$, we obtain
\begin{equation}
\begin{aligned}
\lim_{N \rightarrow \infty} E& \Big[\big|Q_E(s_t, a_t) - Q^\ast(s_t, a_t)\big|\Big]\\ 
\leq & \lim_{N\rightarrow\infty} E\Big[\big|\mu_Q(s_t, a_t)-Q^\ast(s_t, a_t)\big|\Big] \\
& + \beta(n) \cdot \lim_{N\rightarrow\infty} E\big[\sigma_Q(s_t, a_t)\big] \\
= & \; \beta_{\min} \cdot \lim_{N \rightarrow \infty} E\big[\sigma_Q(s_t, a_t)\big].
\end{aligned}
\end{equation}
Because each critic satisfies $\operatorname{Var}\big[Q_{\theta_k}(s_t, a_t)\big] \leq \sigma^2$ for all $k$, the expected standard deviation is bounded, and therefore
\begin{equation}
\lim_{N \to \infty} \mathbb{E}\Big[\big|Q_E(s_t, a_t) - Q^\ast(s_t, a_t)\big|\Big] \leq \beta_{\min} \cdot \sigma.
\end{equation}
In particular, when $\beta_{\min} \to 0$,
\begin{equation}
\lim_{N \to \infty} \mathbb{E}\Big[\big|Q_E(s_t, a_t) - Q^\ast(s_t, a_t)\big|\Big] \leq 0,
\end{equation}
which implies 
\begin{equation}
\mathbb{E}\Big[\big|Q_E(s_t, a_t) - Q^\ast(s_t, a_t)\big|\Big] \xrightarrow{N\to\infty} 0.
\end{equation}
Hence, the ensemble estimator $Q_E(s_t, a_t)$ converges in probability to the true value function $Q^\ast(s_t, a_t)$ as the number of training iterations grows to $\infty$ and $\beta_{\min}$ tends to zero.
\end{proof}

\section{Experimental Results}
\label{experiments}

To comprehensively evaluate our method, we conduct experiments centered around the following research questions:
%\begin{itemize}[leftmargin=2em]
\begin{itemize}
    \item[Q1] Can UACER successfully achieve more robust (see Figs. \ref{comparison_with_baselines}, \ref{heatmap} and Table \ref{vs_worst_adversary}), stable (see Table \ref{stability}), and efficient (see Fig. \ref{comparison_with_baselines}) adversarial RL compared with existing methods?  
    \item[Q2] How crucial is each proposed component in UACER for improving RL performance (see Fig. \ref{ablation})?
    \item[Q3] Does the optimistic design of TDU provide measurable benefits in robustness and convergence compared to conventional pessimistic value estimation under adversarial conditions (see Fig. \ref{pessimistic})?
    \item[Q4] How does critic ensemble size $K$ affect the performance (see Fig. \ref{k_analysis})?
    \item[Q5] Can UACER be universally applicable to other existing robust adversarial RL algorithms (see Fig. \ref{scalability})?
\end{itemize}

\subsection{Environments}
\begin{figure*}[!t]
    \centering
    \subfigure[]
        {\includegraphics[width=0.2\linewidth]{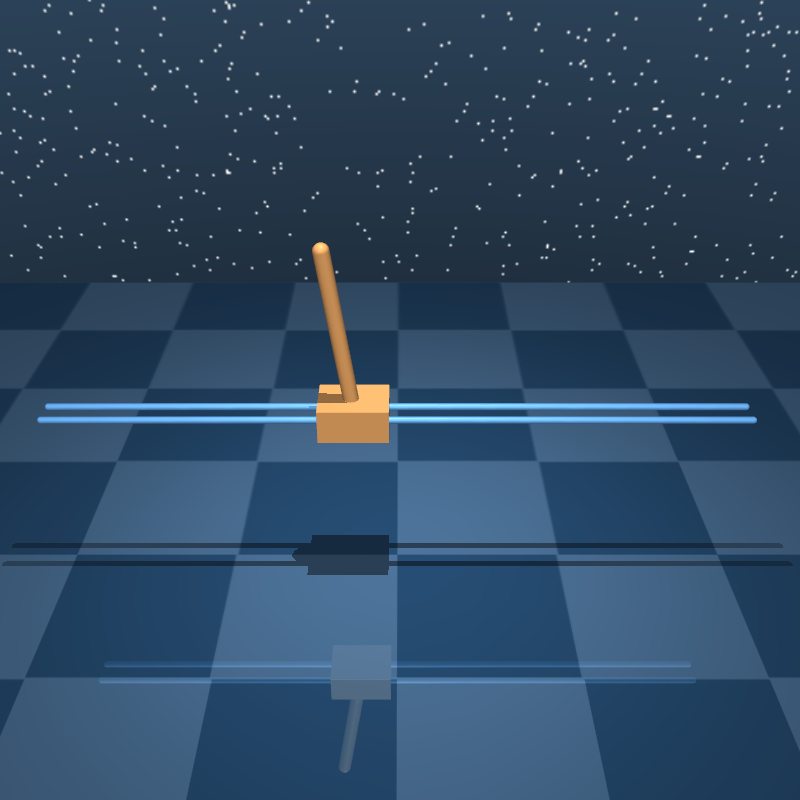}}
    \hspace{.3cm}
    \subfigure[]
        {\includegraphics[width=0.2\linewidth]{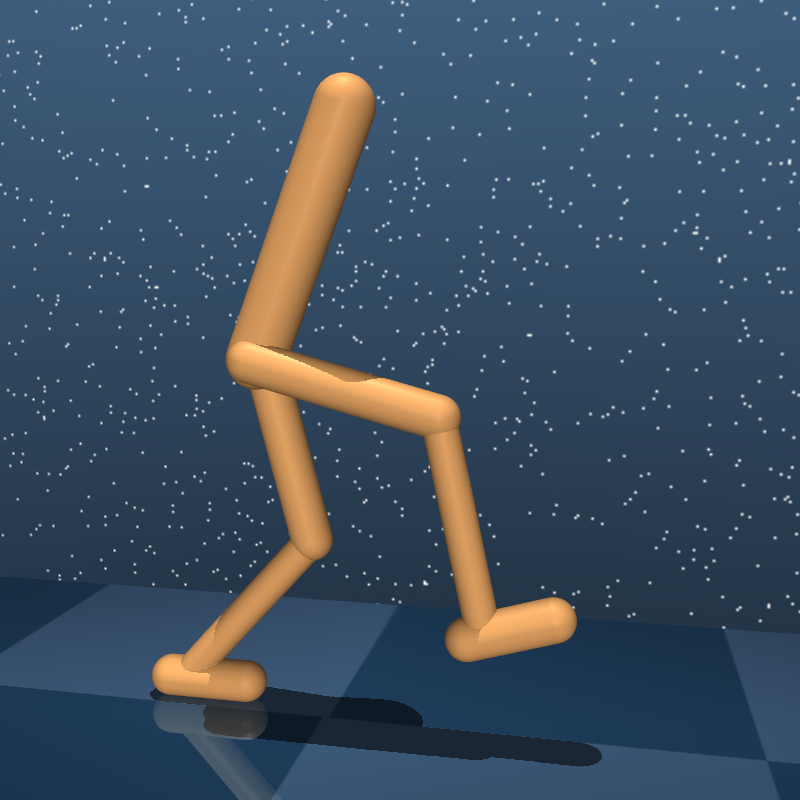}}
    \hspace{.3cm}
    \subfigure[]
        {\includegraphics[width=0.2\linewidth]{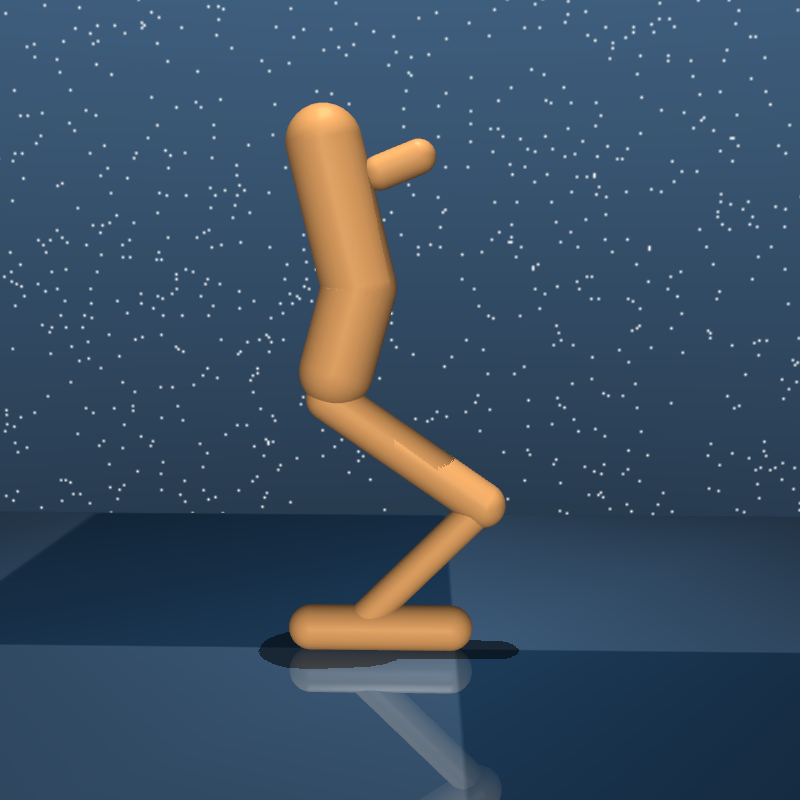}}
    \hspace{.3cm}
    \subfigure[]
        {\includegraphics[width=0.2\linewidth]{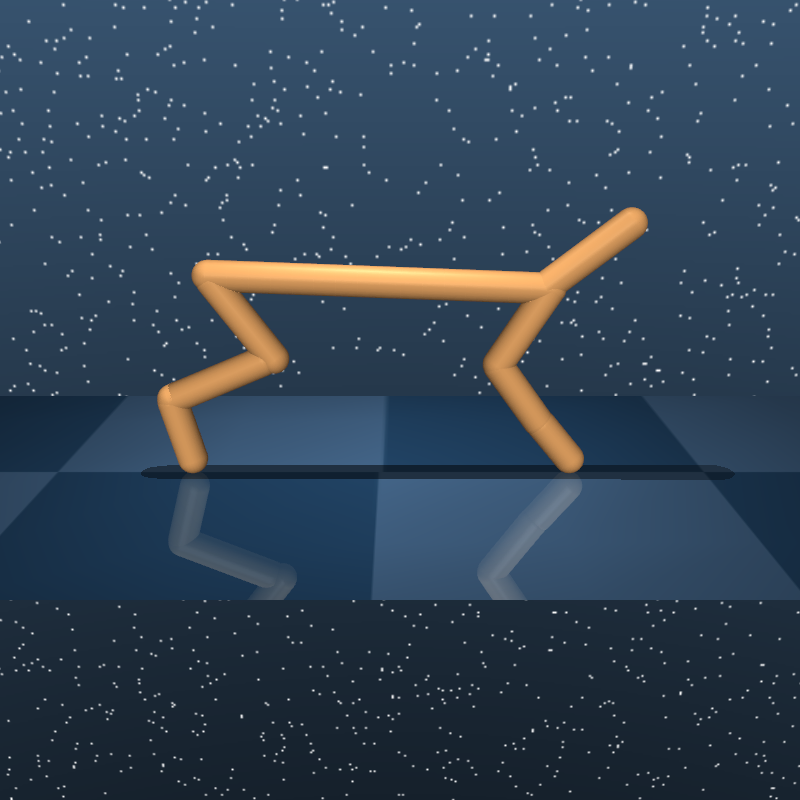}}
    \hspace{.3cm}
    \caption{Representative challenging MuJoCo control tasks from the DeepMind Control Suite including: (a) Cartpole-Swingup Sparse, $|\mathcal{S}|=4$, $|\mathcal{A}|=1$; (b) Walker-Run, $|\mathcal{S}|=18$, $|\mathcal{A}|=6$; (c) Hopper-Hop, $|\mathcal{S}|=14$, $|\mathcal{A}|=4$; (d) Cheetah-Run, $|\mathcal{S}|=18$, $|\mathcal{A}|=6$.}
    \label{environments}
\end{figure*}

We evaluate UACER on several challenging MuJoCo control tasks \cite{todorov2012mujoco} from the DeepMind Control Suite \cite{tunyasuvunakool2020dm_control}, including one sparse-reward task,
{\tt Cartpole-Swingup Sparse}, and three complex locomotion tasks, {\tt Walker-Run}, {\tt Hopper-Hop}, and {\tt Cheetah-Run}.
As shown in Fig. \ref{environments}, each task name is followed by two integers that specify the dimensions of the state and action spaces.
Following established adversarial training protocols \cite{pinto2017robust,DBLP:conf/nips/KamalarubanHHRS20,DBLP:conf/iclr/ReddiT0CD24}, we instantiate two agents: a protagonist, which learns an optimal policy to accomplish the task, and an adversary, which applies external perturbation forces to disrupt the protagonist's performance, thereby eliciting robust behavior.
Both agents observe the standard environmental states as input. The action space of the protagonist coincides with the original action space of the environment, whereas the adversary’s action space is defined as a bounded, multi-dimensional disturbance force.
For each environment, the adversary's perturbation intensity is carefully tuned to be sufficiently large to beget agent robustness and generalization while still posing a meaningful challenge to the protagonist.
The MuJoCo environments follow the same configuration as in QARL \cite{DBLP:conf/iclr/ReddiT0CD24}, with the magnitude of the adversarial force customized for each task.
To prevent catastrophic policy collapse, the maximum adversarial force is explicitly capped in each environment. All environment-specific adversary parameters are listed in Table \ref{environment_parameters}.
% Note that the adversary action spaces reported in here do not apply to baselines CAT \cite{DBLP:conf/ijcnn/ShengZDKCZ22} and MixedNE-LD \cite{DBLP:conf/nips/KamalarubanHHRS20}, as in these methods the protagonist and adversary share the same action space.

\begin{table}[!t]
\centering
\setlength{\tabcolsep}{1.0mm}
\renewcommand\arraystretch{1.2}
\caption{Environment-specific parameters for the adversary.}
\label{environment_parameters}
\begin{tabular}{c|c|c|c}
\toprule
Env.      & \begin{tabular}[c]{@{}c@{}}Adversary\\ Max Force\end{tabular} & \begin{tabular}[c]{@{}c@{}}Performance\\ Lower Bound\end{tabular} & \begin{tabular}[c]{@{}c@{}}Adversary Action\\ Description ($|\mathcal{A}_a|$)\end{tabular} \\
\midrule
Cartpole & 0.005 & 10 & 2D force on the pole (2) \\
Walker & 1.0 & 10 & 2D forces on the feet (4) \\
Hopper & 1.0 & 5 & 2D forces on the foot \& torso (4) \\
Cheetah & 1.0 & 40 & 2D forces on the feet \& torso (6) \\
\bottomrule
\end{tabular}
\end{table}

\subsection{Baselines} 
We evaluate our approach in comparison to the classical robust adversarial RL method RARL \cite{pinto2017robust} and several recently proposed improvements addressing RARL's saddle point optimization challenges, including MixedNE-LD \cite{DBLP:conf/nips/KamalarubanHHRS20}, which utilizes Langevin dynamics to escape local optima, and Curriculum Adversarial Training (CAT) \cite{DBLP:conf/ijcnn/ShengZDKCZ22} featuring dynamic adversary strength adjustment via a hand-designed curriculum.
Additionally, we compare against the state-of-the-art QARL algorithm \cite{DBLP:conf/iclr/ReddiT0CD24}, which synergistically combines entropy regularization with quantal response equilibrium while employing temperature-based curriculum learning to regulate adversary rationality.
For comprehensive assessment, standard SAC \cite{haarnoja2018soft} serves as a non-adversarial baseline representing the underlying RL framework.

Unless specified otherwise, UACER implementations in the following build upon the QARL framework.
For each comparison method, we conduct training over five random seeds $\{0,1,2,3,4\}$ across all environments, with $200$ alternating iterations per seed. All reported results represent the average performance over these five runs.
Complete hyperparameter configurations for all algorithms are provided in {\em Appendix}.
% supplementary materials.

All experiments were run on a workstation equipped with an NVIDIA TITAN Xp GPU (12 GB), 15 GB of system memory, and an Intel\textsuperscript{\textregistered} Xeon\textsuperscript{\textregistered} E5-2680 v4 CPU. The implementations of the algorithms are based on the publicly available codebase of QARL \cite{DBLP:conf/iclr/ReddiT0CD24}, which itself is built on the MushroomRL library \cite{d2021mushroomrl}. The same library is also used to implement the agents and to construct the adversarial environment wrappers.

\subsection{Evaluation Protocols}
To comprehensively assess the effectiveness of the proposed method, we evaluate the learned protagonist from multiple complementary perspectives during and after training. In particular, our evaluation focuses on three key dimensions: the agent’s ability to withstand disturbances, the stability of its learning dynamics, and its final robustness under adversarial conditions.
First, to evaluate disturbance resistance during training, we conduct a {\em robustness evaluation} every five alternating iterations. For each environment, we sweep through a predefined range of environmental parameter variations (e.g., mass and friction coefficient) in the absence of the adversary, and then report the average return across all parameter configurations as the robustness metric.
This procedure introduces graded disturbances via variations in dynamic parameters, offering a thorough test of robustness while simultaneously assessing generalization capability under domain shift conditions.
Concurrently, we quantify {\em training stability} by computing the percentage decrease in the protagonist's performance between consecutive (i.e., $i$ and $i+1$) robustness evaluations and averaging these decreases over the entire training process.
Finally, following the practice in \cite{DBLP:conf/iclr/ReddiT0CD24}, we evaluate the {\em final adversarial performance} of the protagonist against the worst adversary in a minimax sense by fixing the protagonist obtained at the end of training and subsequently training a dedicated adversary against it, and then using that adversary for testing.

\begin{figure*}[!t]
\centering
\includegraphics[width=.98\linewidth]{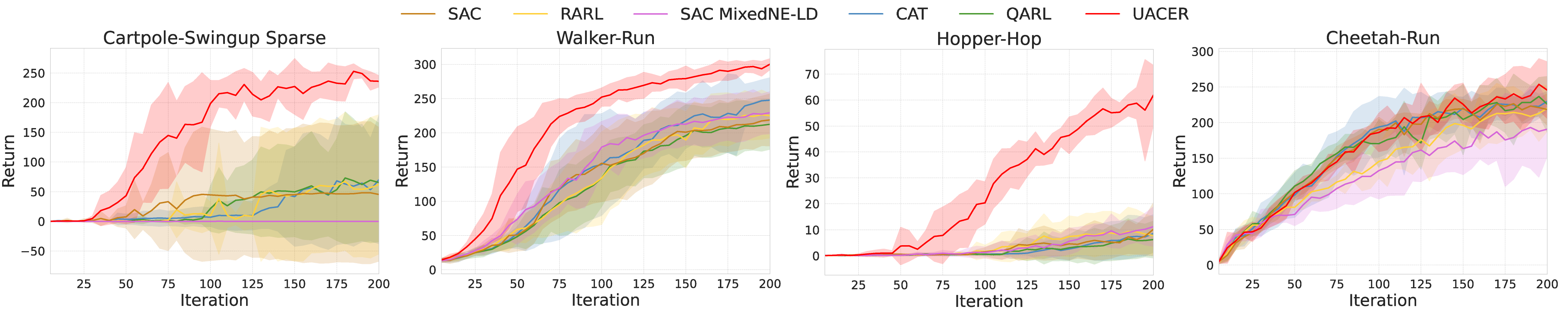}
\caption{Robustness evaluation curves of UACER and baseline methods on MuJoCo swingup and locomotion tasks.}
\label{comparison_with_baselines}
\end{figure*}

\begin{figure*}[!t]
\centering
\includegraphics[width=.98\linewidth]{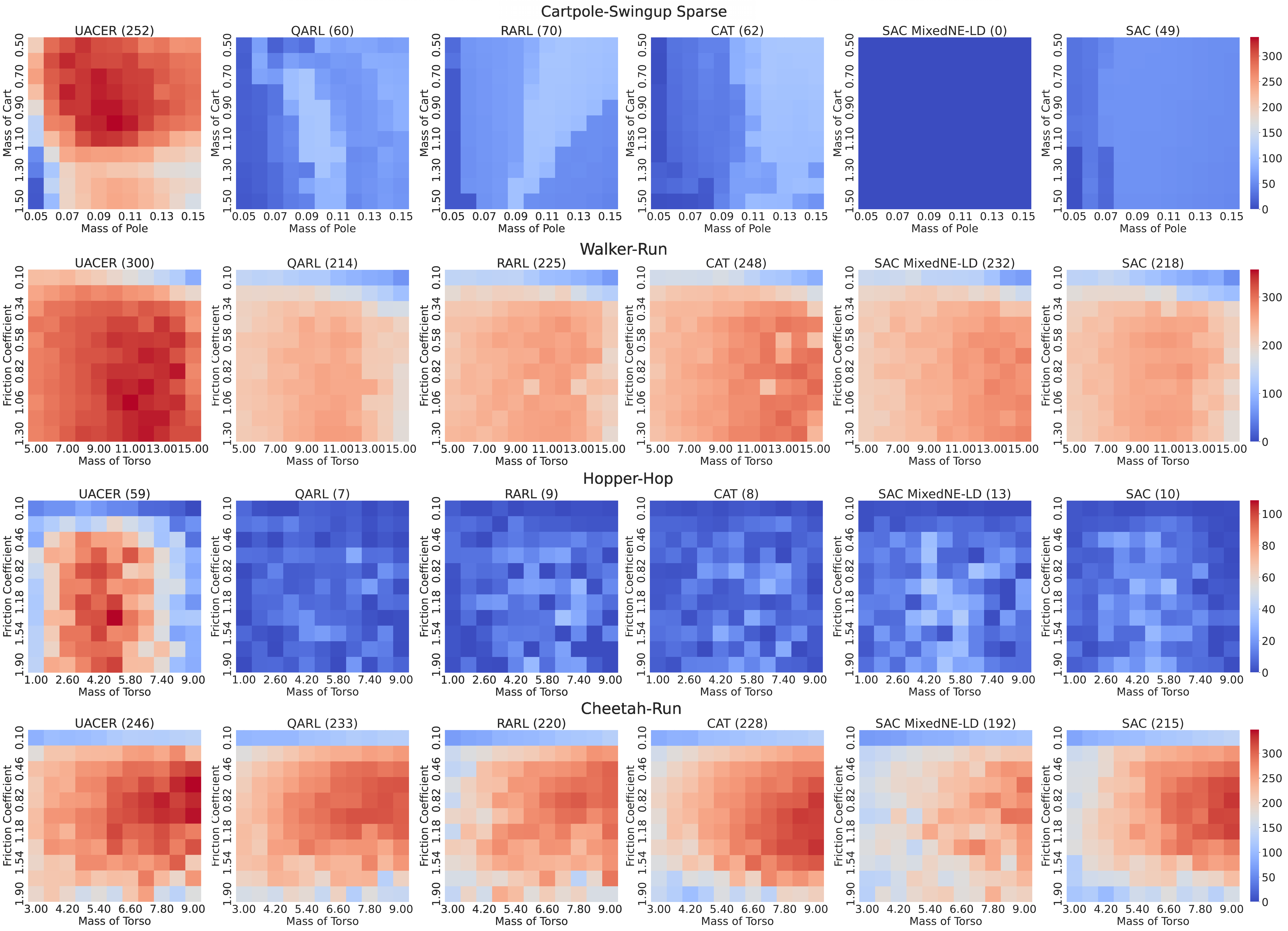}
\caption{Heatmap of final robustness evaluation for UACER and baseline methods on MuJoCo swingup and locomotion tasks, evaluated at the end of training. Each heatmap illustrates the performance achieved for different environmental properties, which are described on the x-y axes.}
\label{heatmap}
\end{figure*}

\subsection{Results} 

To examine the effectiveness of our proposed method (in response to Q1), we conduct experiments on the aforementioned MuJoCo control tasks with UACER and all baseline methods. The experimental results and corresponding analyses are presented below.

\paragraph{Robustness and Convergence}
As demonstrated by the robustness evaluation results in Fig. \ref{comparison_with_baselines}, UACER achieves substantially faster convergence compared to all baseline methods across most environments, effectively mitigating the long-standing challenge of slow convergence in robust adversarial RL. 
This framework consistently attains superior robustness scores in all tested tasks, reflecting both exceptional perturbed resilience and generalization ability. 
Notably, in more complex and sparse-reward environments such as {\tt Hopper-Hop} and {\tt Cartpole-Swingup Sparse}, the baselines exhibit noticeably poorer average performance, as the task complexity and sparse reward setting enable the adversary to easily hinder the protagonist’s policy learning. In contrast, UACER achieves more statistically pronounced performance improvements over competing approaches. 
The key to this success lies in its two core design aspects: the parallel ensemble of value networks reduces Q-value estimation variance and stabilizes policy updates under adversarial perturbations; at the same time, the time-decaying uncertainty-weighted aggregation mechanism adaptively balances exploration and exploitation. By encouraging thorough exploration early in training, it enables the agent to quickly adapt to perturbed environments while maintaining stability throughout the learning process.

Moreover, we visualize the robustness of the trained protagonist to different test conditions upon training completion, as illustrated in Fig. \ref{heatmap}. Each heatmap depicts the performance obtained under varying properties of the environment. 
These results further validate UACER's capability to handle diverse environmental dynamics and also indirectly indicate that the observed comprehensive robustness gains stem from holistic adaptation to various environmental properties, rather than overfitting to any specific environmental dynamics.

\begin{table*}[!t]
\centering
\caption{Training stability analysis of UACER and baseline methods on MuJoCo swingup and locomotion tasks (based on the results in Fig. \ref{comparison_with_baselines}). 
Performance degradation (percentage decrease, $\%$) is reported as the mean $\pm$ standard error over five independent\\ trials. Values shown in \textbf{bold} and \underline{underlined} indicate the best and second-best results, respectively.}
\label{stability}
\setlength{\tabcolsep}{3mm}
\renewcommand\arraystretch{1.2}
\begin{tabular}{c | c | c | c | c}
  \toprule
  Method           & Cartpole-Swingup Sparse        & Walker-Run                       & Hopper-Hop                       & Cheetah-Run          \\
  \midrule
  SAC   & $60.504\pm53.811$ & $3.353\pm1.862$ & $33.367\pm8.704$ & \bm{$4.520\pm4.724$} \\
  RARL             & $56.423\pm40.918$              &  $3.935\pm2.105$     & \underline{$30.215\pm11.733$}    & $6.943\pm2.906$      \\
  SAC MixedNE-LD   & $88.317\pm4.567$                &  \underline{$3.151\pm1.376$}                & $36.220\pm14.331$                 & $6.281\pm2.281$      \\
  CAT              & $56.902\pm51.160$              &  $4.743\pm2.955$                & $32.276\pm12.829$                & \underline{$5.353\pm3.882$} \\
  QARL             & \underline{$35.714\pm28.414$}         &  $4.527\pm0.847$                & $35.581\pm6.025$                 & $8.692\pm4.927$      \\
  UACER            & \bm{$17.352\pm9.593$}   & \bm{$1.659\pm0.604$}             & \bm{$27.938\pm8.474$}            & $7.665\pm3.515$      \\
  \bottomrule

\end{tabular}
\end{table*}

\begin{table*}[!t]
\centering
\caption{Final adversarial performance of UACER and baseline methods on MuJoCo swingup and locomotion tasks, evaluated at the\\ end of training against an adversary trained against the frozen trained protagonist.
Performance metric (return) is\\ reported as the mean $\pm$ standard error over five independent trials. Values shown in \textbf{bold} and \underline{underlined} indicate\\ the best and second-best results, respectively.
Notably, SAC is omitted as it lacks adversarial components.}
\label{vs_worst_adversary}
\setlength{\tabcolsep}{3.0mm}
\renewcommand\arraystretch{1.2}
\begin{tabular}{c | c | c | c | c}
  \toprule
  Method           & Cartpole-Swingup Sparse          & Walker-Run                      & Hopper-Hop                      & Cheetah-Run \\
  \midrule
  RARL             & $107.420\pm182.710$              & \underline{$257.679\pm50.298$}  & $14.267\pm15.725$               & $268.101\pm52.009$  \\
  SAC MixedNE-LD   & $0.000\pm0.000$                  & $233.273\pm41.541$              & \underline{$15.222\pm14.110$}   & $227.447\pm54.054$  \\
  CAT              & $0.000\pm0.000$                  & $229.088\pm23.850$              & $9.514\pm10.039$                & $192.526\pm14.081$  \\
  QARL             & \underline{$111.420\pm189.457$}  & $253.536\pm46.556$              & $9.386\pm8.461$                 & \underline{$292.861\pm42.095$} \\
  UACER            & \bm{$334.770\pm6.886$}          & \bm{$341.995\pm8.480$}          & \bm{$96.564\pm17.479$}          & \bm{$312.594\pm30.422$}  \\
  \bottomrule
\end{tabular}
\end{table*}

\paragraph{Training Stability}
We evaluate the training stability of UACER from two complementary perspectives: the performance variability across different random seeds and the magnitude of return degradation fluctuations during training.
As illustrated in Fig. \ref{comparison_with_baselines}, UACER exhibits the smallest variance in performance curves across different random seeds, as evident from the narrowest shaded regions, indicating minimal sensitivity to random initialization.
This enhanced stability to initialization can be attributed to the critic ensemble framework, which aggregates Q-value predictions from multiple parallel critics to produce more accurate and stable value estimates, thereby mitigating divergent updates caused by different initialization conditions.
Furthermore, the training stability metrics derived from the results in Fig. \ref{comparison_with_baselines} are summarized in Table \ref{stability}, where performance degradation (percentage decrease, $\%$) is reported as the mean $\pm$ standard error over five independent trials. Values shown in \textbf{bold} and \underline{underlined} indicate the best and second-best results, respectively. 
The results demonstrate that UACER achieves the highest training stability in the first three environments, while maintaining performance comparable to the baselines in the last environment (i.e., {\tt Cheetah-Run}). These findings further validate the effectiveness of UACER in alleviating training fluctuations induced by adversarial disturbances in robust adversarial RL settings.

\paragraph{Worst-Case Adversarial Performance}
In addition to the aforementioned comprehensive robustness evaluation under varying environmental parameters, we further follow the minimax principle to examine UACER's ability to cope with stronger and previously unseen adversaries. Specifically, after training is completed, we fix the learned protagonist and continue to train an adversary to minimize the protagonist’s performance, and then evaluate the final adversarial performance between the protagonist and this subsequently optimized, strongest adversary.
As quantitatively summarized in Table \ref{vs_worst_adversary}, UACER consistently outperforms the baseline methods across all environments when facing such further optimized adversaries, while also exhibiting relatively low variance across different initialization settings. These empirical results demonstrate its strong capability to withstand extreme adversarial attacks as well as its superior generalization ability in the presence of untrained perturbations.

\begin{figure*}[t]
\centering
\includegraphics[width=.98\linewidth]{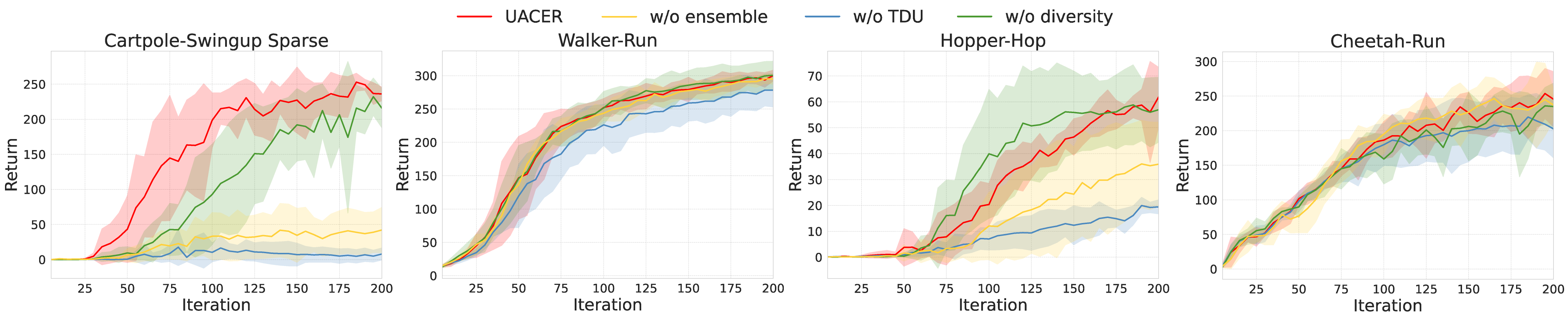}
\caption{Robustness evaluation curves of UACER and its three variants (UACER without ensemble/TDU/diversity) on MuJoCo swingup and locomotion tasks.}
\label{ablation}
\end{figure*}

\begin{figure*}[!t]
\centering
\includegraphics[width=.98\linewidth]{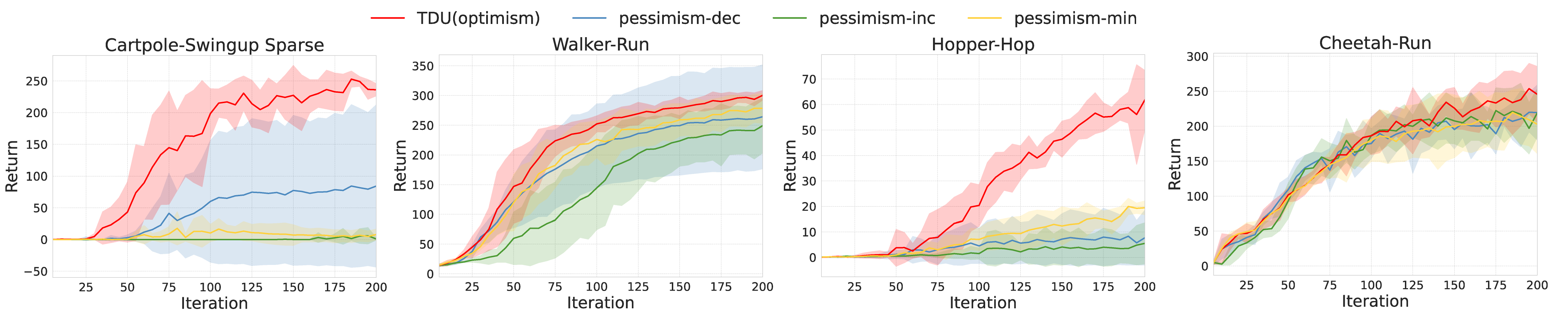}
\caption{Robustness evaluation curves of UACER (optimistic TDU aggregation) and its three pessimistic Q-value estimation variants (UACER with pessimism-dec/pessimism-dec/min aggregation) on MuJoCo swingup and locomotion tasks.}
\label{pessimistic}
\end{figure*}

\subsection{Ablation Study}

In response to Q2, we conduct a systematic ablation study to investigate the individual contributions of UACER's core components, including the critic ensemble, the TDU aggregation mechanism, and the diversity enhancement strategies.
Specifically, we perform controlled single-variable ablation experiments in robust adversarial RL settings by comparing UACER with the following three algorithmic variants:
(i) {\em w/o ensemble}, which replaces the $K$-critic ensemble with the standard two-critic architecture used in the base RL algorithm SAC;
(ii) {\em w/o TDU}, which substitutes the TDU aggregation function with minimum-value selection strategy adopted in SAC; 
(iii) {\em w/o diversity}, which removes the diversity enhancement strategies from UACER.

The results presented in Fig. \ref{ablation} indicate that removing any individual component from UACER leads to varying degrees of performance degradation. 
Specifically, when the diversity enhancement strategy is omitted (``w/o diversity"), the critic networks tend to converge toward similar update patterns, which weakens the ensemble framework’s ability to capture uncertainty under adversarial perturbations. Consequently, the algorithm exhibits noticeably poorer average performance in certain environments, such as {\tt Cartpole-Swingup Sparse} and {\tt Hopper-Hop}.
More critically, removing either the critic ensemble (``w/o ensemble") or the TDU aggregation mechanism (``w/o TDU") results in substantial declines in both robustness and convergence speed across most tasks, particularly in challenging environments characterized by high complexity and sparse rewards.
Furthermore, compared with the other two components, eliminating the TDU mechanism has the most pronounced negative impact on performance across all environments, highlighting the crucial role of the proposed uncertainty-adaptive TDU aggregation mechanism.

In contrast, the complete UACER framework, which integrates all components, consistently outperforms its ablated variants. 
These findings suggest that in environments subject to persistent adversarial disturbances, no single component alone is sufficient to achieve comprehensive robustness across all tasks. 
Instead, the synergistic integration of all components is essential: the critic ensemble provides a solid foundation for stable value estimation; the TDU mechanism effectively converts base model prediction uncertainty into adaptive and stable policy update signals; and, building upon this foundation, the diversity-enhancing network initialization and architectural design further improve training stability and resistance to adversarial perturbations.

\subsection{Analysis}

\begin{figure*}[!t]
    \centering
    \includegraphics[width=.98\linewidth]{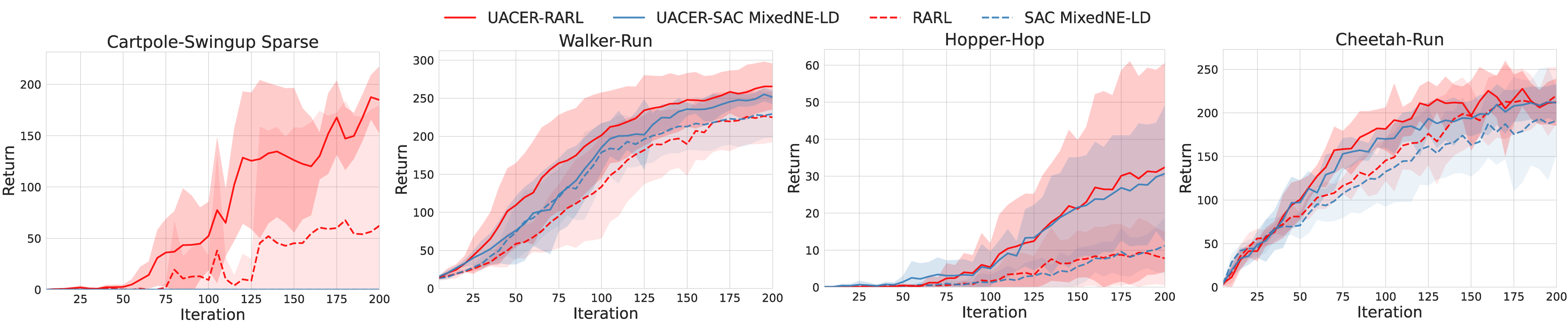}
    \caption{Robustness evaluation curves of UACER combined with other existing robust adversarial RL algorithms including RARL and SAC MixedNE-LD on MuJoCo swingup and locomotion tasks.}
    \label{scalability}
\end{figure*}

\begin{figure}[!t]
    \centering
    \subfigure[\label{k_analysis_walker}]
        {\includegraphics[width=0.98\linewidth]{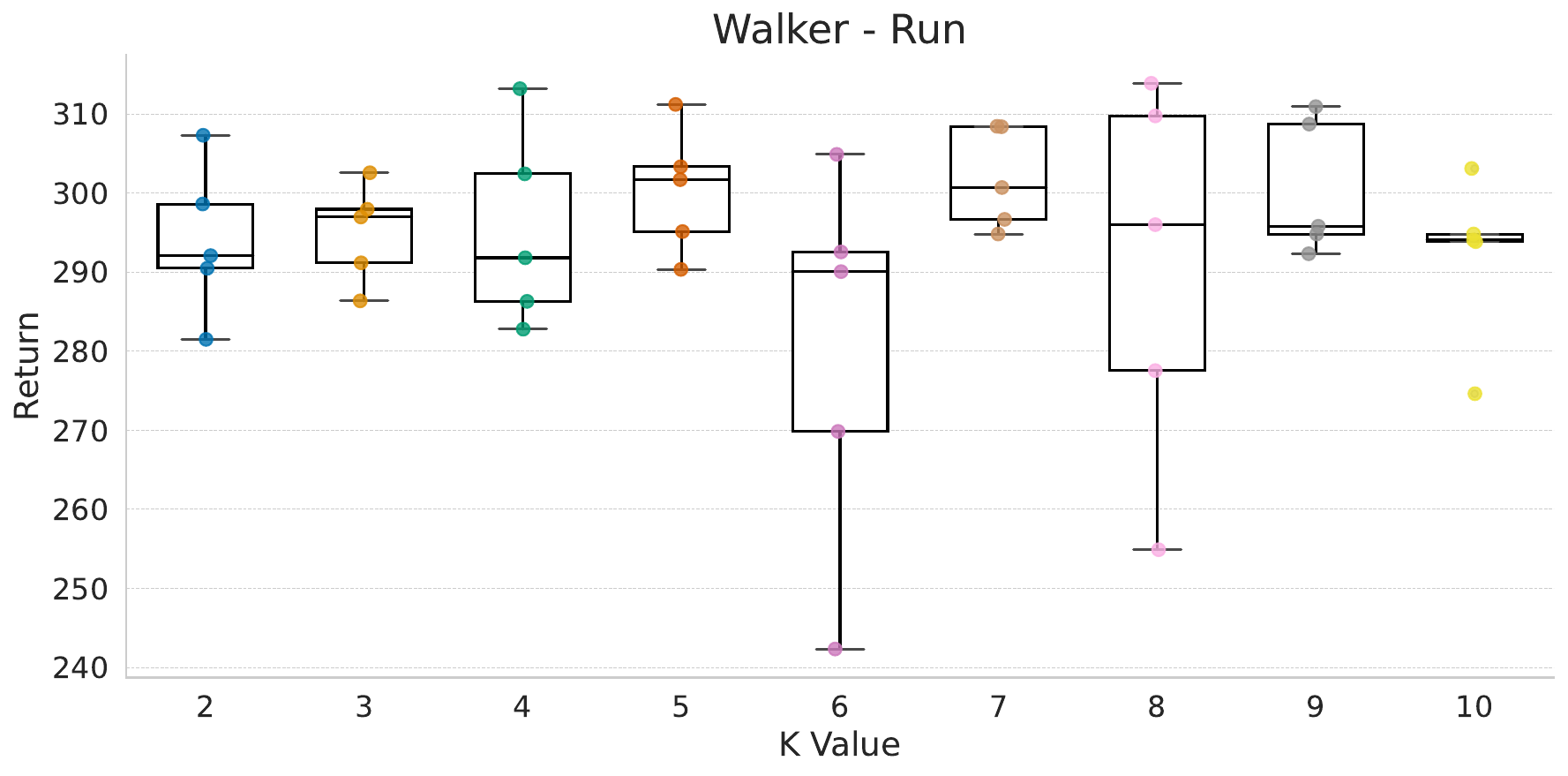}}
    \subfigure[\label{k_analysis_hopper}]
        {\includegraphics[width=0.98\linewidth]{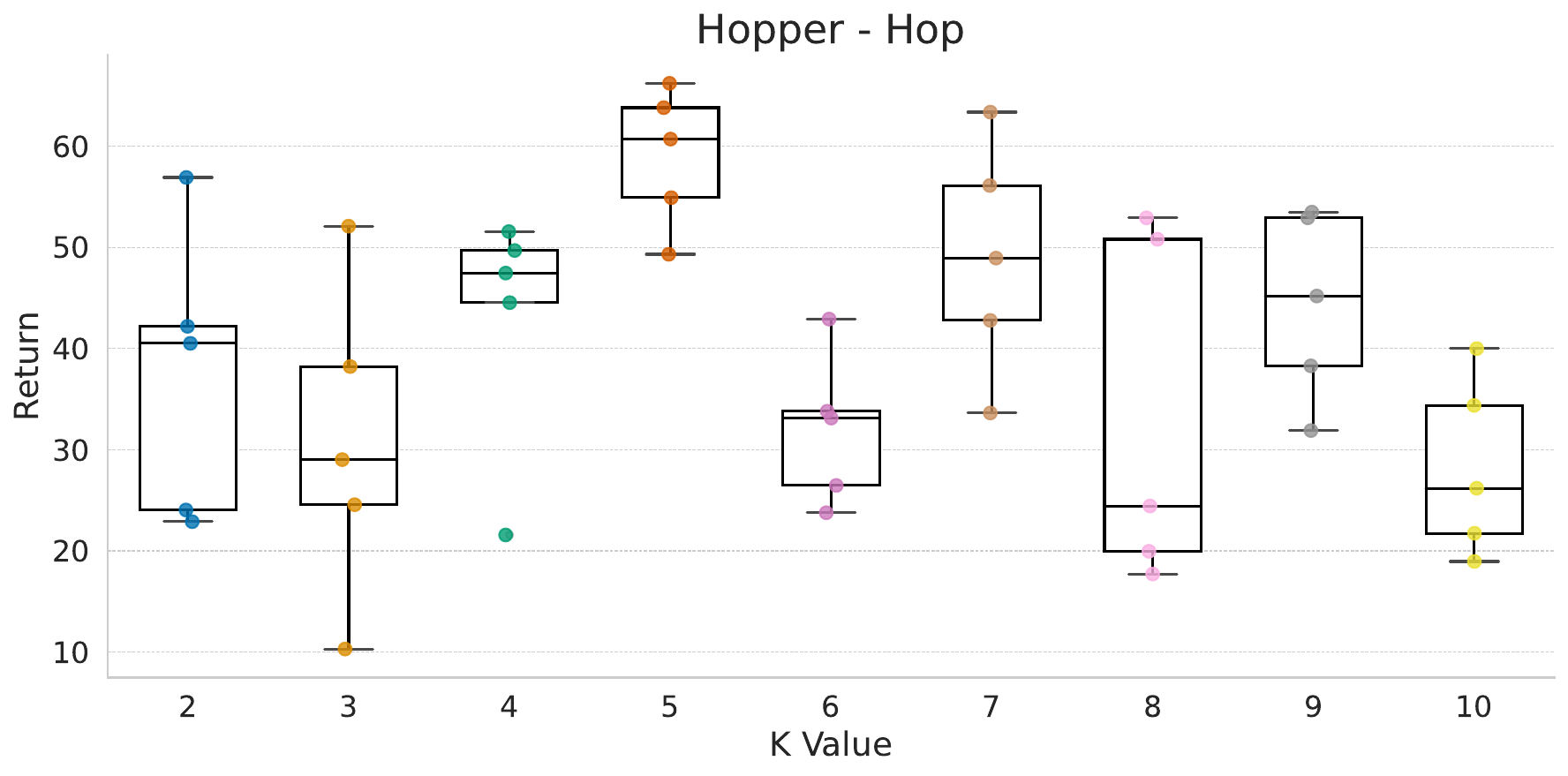}}
    \caption{Robustness evaluation of UACER with varying critic ensemble sizes $K$ on the {\tt Walker-Run} and {\tt Hopper-Hop} tasks.}
    \label{k_analysis}
\end{figure}

\paragraph{Superiority of TDU's Optimistic Aggregation in Adversarial RL}
The principle of optimistic estimation in TDU differs fundamentally from the widely adopted pessimistic strategies that aim to mitigate overestimation bias. Conventional approaches \cite{fujimoto2018addressing, haarnoja2018soft, DBLP:conf/iclr/LanPFW20} typically implement pessimism by selecting the lower of two Q-network outputs, such as min$(Q_1, Q_2)$, to suppress excessively high estimates.
However, in uncertain or adversarially disturbed environments, such pessimistic mechanisms tend to induce over-conservatism, which may incorrectly suppress truly valuable actions, thereby harming the agent's learning efficiency and overall performance.
In contrast, UACER employs a controlled optimistic estimator that does not indiscriminately inflate Q-values. Instead, the estimator restricts the degree of optimism within an upper-confidence bound, ensuring that the Q-value estimates remain appropriately optimistic while strictly adhering to theoretically grounded confidence bounds. This design achieves a balance between training stability and effective exploration.

To evaluate the superiority of TDU's optimistic aggregation in adversarial RL  (in response to Q3), we compare it with three pessimistic variants derived from Eq. \eqref{TDU_func}:
(i) {\em pessimism-dec}: the mean {\em minus} the uncertainty term, with $\beta(n)$ decaying from 1 to 0;
(ii) {\em pessimism-inc}: the mean {\em minus} the uncertainty term, with $\beta(n)$ increasing from 0 to 1;
(iii) {\em pessimism-min}: the minimum among the $K$ critic outputs.
As demonstrated in Fig. \ref{pessimistic}, TDU consistently delivers superior performance across all tasks. This advantage arises because, in non-stationary adversarial environments, the protagonist already struggles to learn appropriate responses to perturbations, and pessimistic estimation further constrains its Q-value prediction capacity under such disturbances.
By contrast, TDU's optimistic aggregation achieves a more effective balance between bias control and robustness enhancement, thereby alleviating the tendency toward over-conservative policy learning while simultaneously strengthening training stability and final performance.

\paragraph{Effects of Critic Ensemble Size}
The ensemble size $K$, i.e., the number of critic networks, is a key hyperparameter newly introduced in the proposed UACER framework. To analyze its impact (in response to Q4), we conduct controlled experiments in which $K$ is varied from $2$ to $10$ while all other hyperparameters remain fixed.
Fig. \ref{k_analysis} presents box plots illustrating two distinct trends of agent robustness performance with respect to the value of $K$ on the tasks {\tt Walker-Run} and {\tt Hopper-Hop}. 
It can be observed that the agent's performance does not increase monotonically with larger ensemble sizes, contrary to intuitive expectations.
Overall, when $K$ is relatively small (e.g., $K<5$), the limited ensemble diversity is insufficient to fully exploit the stability gains of critic ensemble for Q-value estimation. In this regime, increasing $K$ gradually improves the agent’s robustness performance.
However, once $K$ exceeds a certain threshold (e.g., $K>5$), the performance gains begin to saturate and may even decline to varying degrees. 
A possible explanation for this phenomenon is that, although a larger ensemble size can theoretically enhance diversity and thereby stabilize Q-value estimation, an excessive number of critics may introduce detrimental redundancy and complicate policy optimization.
Consequently, a larger $K$ does not necessarily lead to better performance, and its appropriate value should be determined according to the complexity of the specific task. In this work, balancing the agent's performance across the examined range of $K$ values and the associated computational cost, we select $K=5$ for all experimental validation.

\paragraph{Scalability Analysis}
All previous experiments implemented UACER based on the QARL framework as the foundational robust adversarial RL approach. In response to Q5, this section examines UACER's applicability as a general framework when integrated with other existing robust adversarial RL algorithms. 
Specifically, we migrate UACER to the classic RARL algorithm and its improved extension SAC MixedNE-LD for practical implementation and evaluation.
The learning curves in terms of robustness performance are shown in Fig. \ref{scalability}.
The experimental results confirm that both RARL and SAC MixedNE-LD achieve substantial performance gains after incorporating UACER, showing improved convergence speed, training stability, and robustness to adversarial perturbations.
In essence, UACER is designed as an algorithm-agnostic framework that is compatible with any value-based robust adversarial RL methods.
Regrettably, in the {\tt Cartpole-Swingup Sparse} task, SAC MixedNE-LD exhibited critically poor baseline performance (near-zero returns throughout training), with no discernible improvement after UACER integration. This outcome indicates that UACER's performance improvements remain partly dependent on the base adversarial RL algorithm's intrinsic adaptability to the specific task.

\paragraph{Computational Overhead and Limitations}
A potential limitation of UACER is the additional computational cost introduced by updating multiple critics. Nevertheless, this overhead can be effectively mitigated through parallel execution, which prevents wall-clock training time from rising substantially compared to single-critic baselines. 
By performing critic updates concurrently, the incremental per-iteration computation is effectively absorbed, maintaining overall training efficiency. 
In return, UACER delivers notable improvements in training stability, convergence speed, and robustness against adversarial perturbations. 
Given these performance gains, the elevated computational demand remains acceptable and manageable in practice, representing a favorable trade-off relative to conventional single-critic adversarial RL architectures.

\section{Conclusion}
In this article, we present UACER, a novel robust adversarial RL approach that advances the field through two core contributions: 
(i) a parallel ensemble of critic networks that simultaneously reduces Q-value estimation variance and enhances robustness under adversarial perturbations, and
(ii) a time-varying decay uncertainty Q-value aggregation mechanism that adaptively balances exploration and exploitation through variance-based epistemic uncertainty quantification across critics.  
Supported by theoretical analysis and extensive experimental validation, UACER achieves significant improvements in training stability, convergence speed, and adversarial robustness, effectively addressing critical challenges that hinder the reliable deployment of deep RL systems in real-world applications.
By integrating ensemble-based value estimation with uncertainty-adaptive policy optimization, the proposed framework provides a systematic methodology for developing RL algorithms with provable stability guarantees and consistent performance in complex and uncertain environments.

Future research directions include extending the UACER architecture to partially observable settings, where information constraints introduce additional sources of uncertainty,
and exploring its applications in real-world safety-critical systems, such as robotic manipulation and autonomous driving scenarios, where both robustness and adaptability are essential for safe and reliable operation.

%\section*{Acknowledgments}
%This should be a simple paragraph before the References to thank those individuals and institutions who have supported your work on this article.

\bibliographystyle{IEEEtran}
\bibliography{IEEEtrans}

 

% \end{IEEEbiography}



{\appendix

\section*{Hyperparameter Setting}
\label{hyperparameter_setting}

To ensure fair and valid comparisons, we strictly follow the experimental protocol established by QARL \cite{DBLP:conf/iclr/ReddiT0CD24}, maintaining consistent shared hyperparameter settings across all methods. This practice ensures that performance differences can be attributed to algorithmic distinctions rather than variations in parameter tuning. The hyperparameters used for UACER and each baseline method are provided in Table \ref{hyperparameters}.

\begin{table}[!t]
\centering
\setlength{\tabcolsep}{1.5mm}
\renewcommand\arraystretch{1.3}
\caption{Algorithm hyperparameters.}
\label{hyperparameters}
\begin{tabular}{ll}
\toprule
Hyperparameter   & Value \\
\midrule
\multicolumn{2}{l}{\textit{Shared (ALL)}} \\
\quad \# iterations $N$ & 200 \\
\quad \# episodes per agent per iteration & 5 \\
\quad \# evaluation rollouts per iteration & 10 \\
\quad \# hidden layers & 3 \\
\quad \# hidden units per layer & 256 \\
\quad discount factor $\gamma$ & 0.99 \\
\quad horizon & 500 \\
\midrule
\multicolumn{2}{l}{\textit{UACER}} \\
\quad \# ensemble critic networks $K$ & 5 \\
\quad $\beta_0$ & 0.85 \\
\quad $\beta_{min}$ & 0.15 \\
\quad decay speed $\lambda$ & 3 \\
\midrule
\multicolumn{2}{l}{\textit{Shared (SAC and SAC MixedNE-LD)}} \\
\quad nonlinearity & ReLU \\
\quad critic optimiser & Adam \\
\quad critic learning rate & $3 \times 10^{-4}$ \\
\quad actor learning rate & $1 \times 10^{-4}$ \\
\quad initial replay memory size & $3 \times 10^{3}$ \\
\quad max replay memory size & $1 \times 10^{6}$ \\
\quad warmup transitions & $5 \times 10^{3}$ \\
\quad batch size & 256 \\
\quad target smoothing coefficient & $5 \times 10^{-3}$ \\
\quad target update interval & 1 \\
\quad policy log std bounds & $[-20, 2]$ \\
\quad initial temperature & $5 \times 10^{-3}$ \\
\quad temperature learning rate & $3 \times 10^{-4}$ \\
\quad target entropy & $-\textrm{dim}(\mathcal{A})$ \\
\midrule
\multicolumn{2}{l}{\textit{SAC}} \\
\quad actor optimiser & Adam \\
\midrule
\multicolumn{2}{l}{\textit{SAC MixedNE-LD}} \\
\quad adversary influence & 0.1 \\
\quad actor optimiser & SGLD \\
\quad thermal noise & $10^{-3} \times (1 - 5 \times 10^{-5})^t$ \\
\quad RMSProp parameter & 0.999 \\
\quad RMSProp parameter & $10^{-8}$ \\
\midrule
\multicolumn{2}{l}{\textit{CAT}} \\
\quad curriculum start iteration & $0.2 \times \text{\# iterations}$ \\
\quad curriculum end iteration & $0.8 \times \text{\# iterations}$ \\
\multicolumn{2}{l}{\textit{CAT MAS Adversary}} \\
\quad gradient descent learning rate & 3 \\
\quad gradient descent step limit & 25 \\
\quad gradient descent convergence threshold & $10^{-3}$ \\
\quad disturbance $L^p$-norm & 2 \\
\midrule
\multicolumn{2}{l}{\textit{QARL}} \\
\quad initial gamma distribution concentration & 50 \\
\quad target gamma distribution concentration & 1 \\
\quad fixed gamma distribution rate & 1000 \\
\quad $D_{\text{KL}}$ constraint & 0.5 \\
\quad \# rollouts needed for update & 30 \\
\bottomrule
\end{tabular}
\end{table}

}

% \vfill

\end{document}